\documentclass{article}
\usepackage{ijcai22}

\usepackage{times}
\usepackage{soul}
\usepackage{url}
\usepackage[hidelinks]{hyperref}
\usepackage[utf8]{inputenc}
\usepackage[small]{caption}
\usepackage{graphicx}
\usepackage{amsmath}
\usepackage{amsthm}
\usepackage{amsfonts,amssymb}
\usepackage{booktabs}
\usepackage{algorithm}
\usepackage{algorithmic}
\usepackage{xcolor}
\usepackage{multirow}
\usepackage{colortbl}
\usepackage{multirow}
\usepackage{subfigure}
\definecolor{LightCyan}{rgb}{0.88,1,1}
\definecolor{LightYellow}{rgb}{1,0.88,1}
\definecolor{LightRed}{rgb}{1,1,0.88}
\definecolor{LightGray}{gray}{0.8}
\usepackage{placeins}
\long\def\/*#1*/{}

\title{Transformer for Graphs: An Overview from Architecture Perspective}

\author{
Erxue Min$^1$\footnote{Email: erxue.min@manchester.ac.uk}\and
Runfa Chen$^3$\and
Yatao Bian$^2$\and
Tingyang Xu$^2$\and
Kangfei Zhao$^2$\and
Wenbing Huang$^4$\and
Peilin Zhao$^2$\and
Junzhou Huang$^5$\and
Sophia Ananiadou$^1$\and
Yu Rong$^{2}$\footnote{Corresponding Author: yu.rong@hotmail.com}
\affiliations
$^1$National Centre for Text Mining, Department of Computer Science, The University of Manchester\\
$^2$Tencent AI lab\\
$^3$Department of Computer Science and Technology, Tsinghua University\\
$^4$Institute for AI Industry Research (AIR), Tsinghua University\\
$^5$Department of Computer Science and Engineering, University of Texas at Arlington
}

 \usepackage{cleveref} 
 \crefname{section}{Section}{Sections}
 \crefname{theorem}{Theorem}{Theorems}
 \crefname{lemma}{Lemma}{Lemmas}
 \crefname{equation}{Equation}{Equations}
 \crefname{proposition}{Proposition}{Propositions}
 \crefname{claim}{Claim}{Claims}
\crefname{appendix}{Appendix}{Appendices}
   \crefname{algorithm}{Algorithm}{Algorithms}
 \crefname{figure}{Figure}{Figures}
 \crefname{table}{Table}{Tables}
 \crefname{remark}{Remark}{Remarks}
 \crefname{definition}{Definition}{Definitions}
 \crefname{equatinon}{Equation}{Equations}
 \crefname{corollary}{Corollary}{Corollaries}

\allowdisplaybreaks

\usepackage{thmtools}
\usepackage{thm-restate}

\let \oldtextcircled \textcircled
\renewcommand{\textcircled}[1]{\oldtextcircled{\footnotesize #1}}

\usepackage{enumitem}
\setlist[itemize]{leftmargin=9mm}

\def \b{\mathbf{b}}

\def \x{\mathbf{x}}

\def \z{\mathbf{z}}

\def \BO{\mathbf{O}}
\def \BK{\mathbf{K}}
\def \BA{\mathbf{A}}
\def \BB{\mathbf{B}}
\def \BX{\mathbf{X}}

\def \BI{\mathbf{I}}

\def \BD{\mathbf{D}}
\def \BE{\mathbf{E}}

\def \BM{\mathbf{M}}

\def \BP{\mathbf{P}}
\def \BQ{\mathbf{Q}}

\def \BU{\mathbf{U}}
\def \BV{\mathbf{V}}
\def \BW{\mathbf{W}}
\def \BZ{\mathbf{Z}}

\usepackage{mathtools}

\renewcommand{\mid}{|}

\begin{document}

\maketitle

\begin{abstract}
Recently, Transformer model, which has achieved great success in many artificial intelligence fields, has demonstrated its great potential in modeling graph-structured data. 
Till now, a great variety of Transformers has been proposed to adapt to the graph-structured data. However, a comprehensive literature review and systematical evaluation of these Transformer variants for graphs are still unavailable. 
It's imperative to sort out the existing Transformer models for graphs and systematically investigate their effectiveness on various graph tasks. In this survey, we provide a comprehensive review of various Graph Transformer models from the architectural design perspective. We first disassemble the existing models and conclude three typical ways to incorporate the graph information into the vanilla Transformer: 1) GNNs as Auxiliary Modules, 2) Improved Positional Embedding from Graphs, and 3) Improved Attention Matrix from Graphs. Furthermore, we implement the representative components in three groups and conduct a comprehensive comparison on various kinds of famous graph data benchmarks to investigate the real performance gain of each component.  Our experiments confirm the benefits of current graph-specific modules on Transformer and reveal their advantages on different kinds of graph tasks.


\end{abstract}

\section{Introduction}



Graph is a kind of data structure that structurally depicts a set of objects (nodes) with their relationships (edges). As a unique non-Euclidean data structure, graph analysis focuses on tasks such as node classification \cite{yang2016revisiting}, link prediction \cite{zhang2018link}, and clustering \cite{Aggarwal2010}. 
Recent research on analyzing graphs using deep learning has attracted more and more attention due to the rich expressive power of deep learning models.
Graph Neural Networks (GNNs) \cite{kipf2016semi}, as a kind of deep learning-based method, are recently becoming a widely-used graph analysis tools due to their convincing performance. 
Most of the current GNNs are based on the Message Passing paradigm \cite{gilmer2017neural}, the expressive power of which is bounded by the Weisfeiler-Lehamn isomorphism hierarchy \cite{maron2019provably,xu2018powerful}. Worse still, as pointed out by \cite{kreuzer2021rethinking},  GNNs suffer from the $\textit{over-smoothing}$ problem due to repeated local aggregation, and the $\textit{over-squashing}$ problem due to the exponential computation cost with the increase of model depth. Several studies \cite{zhao2019pairnorm,rong2019dropedge,xu2018representation} try to address such problems. Nevertheless, none of them seems to be able to eliminate these problems from the Message Passing paradigm.

On the other hand, Transformers and its variants, as a powerful class of models, are playing an important role in various areas including Natural Language Processing (NLP)~\cite{vaswani2017attention}, Computer Vision (CV)~\cite{forsyth2011computer}, Time-series Analysis~\cite{hamilton2020time}, Audio Processing~\cite{purwins2019deep}, etc. Moreover, recent years have witnessed many successful Transformer variants in modeling graphs. These models have achieved competitive or even superior performance against GNNs in many applications, such as Quantum Property Prediction \cite{hu2021ogblsc}, Catalysts Discovery \cite{ocp_dataset} and Recommendation Systems \cite{min2022masked}. 
The literature reviews and systematic evaluations for these Transformer variants are, however, lacking.


This survey gives an overview of the current research progress on incorporating Transformers in graph-structured data.
Concretely, we provide a comprehensive review of over 20 Graph Transformer models from the architectural design perspective. 
We first dismantle the existing models and conclude three typical ways to incorporate the graph information into the vanilla Transformer: \textbf{1) GNNs as Auxiliary Modules}, \emph{i.e.}, directly inject GNNs into Transformer architecture. Specifically, according to the relative position between GNN layers and Transformer layers, existing GNN-Transformer architectures can be categorized into three types: (a) build Transformer blocks on top of GNN blocks, (b) stack GNN blocks and Transformer blocks in each layer, (c) parallel GNN blocks and Transformer blocks in each layer. 
\textbf{2) Improved Positional Embedding from Graphs}, \emph{i.e.}, compress the graph structure into positional embedding vectors and add them to the input before it is fed to the vanilla Transformer model. This graph positional embedding can be derived from the structural information of graphs, such as degree and centrality. 
\textbf{3) Improved Attention Matrices from Graphs}, \emph{i.e.}, inject graph priors into the attention computation via graph bias terms, or restrict a node only attending to local neighbours in the graph, which can be computationally formulated as an attention masking mechanism.

Additionally, in order to investigate the effectiveness of existing models in various kinds of graph tasks, we implement the representative components in three groups and conduct comprehensive ablation studies on six popular graph-based benchmarks to uniformly test the real performance gain of each component. 
Our experiments indicate that: 1). Current models incorporating the graph information can improve the performance of Transformer on both graph-level and node-level tasks. 2).  Utilizing GNN as auxiliary modules and improving attention matrix from graphs generally contributes more performance gains than encoding graphs into positional embeddings. 3). The performance gain on graph-level tasks is more significant than that on node-level tasks. 
4) Different kinds of graph tasks enjoy different group of models.

The rest of this survey is organized as follows. We first review the general vanilla Transformer Architecture in Section~\ref{sec.prelimiary}.  Section~\ref{sec.transformer_on_graphs} summarizes existing works about the Transformer variant on Graphs and systematically categorizes these methods into three groups. In Section~\ref{sec.exp}, comprehensive ablation studies are conducted to verify the effectiveness and compatibility of these proposed models. In Section~\ref{sec.conclusion}, we conclude this survey and discuss several future research directions.



\section{Transformer Architecture}\label{sec.prelimiary}
Transformer~\cite{vaswani2017attention} architecture was first applied to machine translation. 
In the paper, Transformer is introduced as a novel encoder-decoder architecture built with multiple blocks of self-attention. Let $\BX \in \mathbb{R}^{n\times d}$ to be the input of each Transformer layer, where $n$ is number of tokens, $d$ is the dimension of each token, 
then one block layer can be a function $f_\theta: \mathbb{R}^{n \times d} \rightarrow \mathbb{R}^{n \times d}$ with $f_\theta(\BX)=:\BZ$ defined by:
\begin{align}
\mathbf{A}&=\frac{1}{\sqrt{d}} \mathbf{X Q}(\mathbf{X}\mathbf{K})^{\top}, \label{eq:attention1}\\
\widetilde{\BX}&=\operatorname{SoftMax}(\mathbf{A}) (\mathbf{X} \mathbf{V}), \label{eq:attention2}\\
\mathbf{M}&=\operatorname{LayerNorm}_{1}(\widetilde{\BX}\BO+\mathbf{X}), \label{eq:attention3}\\
\mathbf{F}&=\sigma(\mathbf{M W}_1+\mathbf{b}_1) \mathbf{W}_2+\mathbf{b}_2, \label{eq:ffn}\\
\mathbf{Z}&=\operatorname{LayerNorm}_{2}(\mathbf{M}+\mathbf{F}), \label{eq:layernorm}
\end{align}
where Equation \ref{eq:attention1}, Equation \ref{eq:attention2}, and Equation \ref{eq:attention3} are the attention computation; while Equation \ref{eq:ffn} and Equation \ref{eq:layernorm} are the position-wise feed-forward network (FFN) layers.
Here, $\text{Softmax}(\cdot)$ refers to the row-wise softmax function, $\text{LayerNorm}(\cdot)$ refers to layer normalization function \cite{ba2016layer}, and $\sigma$ refers to the activation function. $\BQ,\BK,\BV,\BO \in \mathbb{R}^{d \times d}, \BW_1 \in \mathbb{R}^{d\times d_f},\b_1 \in \mathbb{R}^{d_f},\BW_2 \in \mathbb{R}^{d_f\times d} ,\b_2 \in \mathbb{R}^{d}$ are trainable parameters in the layer.
Furthermore, it is common to consider multiple attention heads to extend the self-attention to Multi-head Self-attention (MHSA). Specifically, $\BQ,\BK,\BV$ are decomposed into $H$ heads with $\BQ^{(h)},\BK^{(h)},\BV^{(h)} \in \mathbb{R}^{d\times d_h}$ with $d=\sum_{h=1}^{H}d_h$, and the matrices $\widetilde{\BX}^{(h)} \in \mathbb{R}^{n \times d_h}$ from attention heads are concatenated to obtain $\widetilde{\BX}$. In this case,
Equation \ref{eq:attention1} and Equation \ref{eq:attention2} respectively become:
\begin{align}
\BA^{(h)}=\frac{1}{\sqrt{d}} \BX \BQ^{(h)}(\BX\BK^{(h)})^{\top}, \\
\widetilde{\BX}=\|_{h=1}^{H}(\operatorname{SoftMax}(\BA^{(h)}) \BX \BV^{(h)}).
\end{align}
The multi-head mechanism enables the model to implicitly learn representation from different aspects.
Apart from the attention mechanism, the paper uses $\operatorname{sine}$ and $\operatorname{cosine}$ functions with different frequencies as positional embedding to distinguish the position of each token in the sequence.

\begin{table}[]
\centering
\caption{A summary of papers that applied Transformers on graph-structured data. \textbf{GA}: GNNs as Auxiliary Modules; \textbf{PE}: Improved Positional Embedding from Graphs; \textbf{AT}: Improved Attention Matrix from Graphs. }
\label{tab_summary}
\small
\setlength\tabcolsep{3pt}{
\resizebox{0.49\textwidth}{!}{
\begin{tabular}{c|ccc|c}
\toprule
\textbf{Method}   &  \textbf{GA} & \textbf{PE} & \textbf{AT}  & \textbf{Code}   \\ 
\midrule
\cite{zhu2019modeling} &  & &\checkmark & \textcolor{blue}{\href{https://github.com/Amazing-J/ structural-transformer}{\checkmark}}\\\hline
\cite{shiv2019novel} &  &\checkmark & &  \\\hline
\cite{wang2019self} &  &\checkmark &\checkmark &\\\hline
U2GNN~\cite{Nguyen2019UGT} &  & &\checkmark &\textcolor{blue}{\href{https://github.com/daiquocnguyen/Graph-Transformer}{\checkmark}}\\\hline
HeGT~\cite{yao2020heterogeneous} &  & &\checkmark  &\textcolor{blue}{\href{https://github.com/QAQ-v/HetGT}{\checkmark}}\\\hline
Graformer~\cite{schmitt2020modeling} &  & &\checkmark &\textcolor{blue}{\href{https://github.com/mnschmit/graformer}{\checkmark}}\\\hline
PLAN~\cite{khoo2020interpretable} & & &\checkmark &\textcolor{blue}{\href{https://github.com/serenaklm/rumor_detection}{\checkmark}}\\\hline
UniMP~\cite{shi2020masked} &  & &\checkmark &\\\hline
GTOS~\cite{cai2020graph} &  & \checkmark&\checkmark &\textcolor{blue}{\href{https://github.com/jcyk/gtos}{\checkmark}}\\\hline
Graph Trans~\cite{dwivedi2020generalization} &  &\checkmark &\checkmark &\textcolor{blue}{\href{https://github.com/graphdeeplearning/graphtransformer}{\checkmark}}\\\hline
Grover~\cite{rong2020self} &  \checkmark & & &\textcolor{blue}{\href{https://github.com/tencent-ailab/grover}{\checkmark}}\\\hline
Graph-BERT~\cite{zhang2020graph} &  \checkmark & \checkmark & &\textcolor{blue}{\href{https://github.com/jwzhanggy/Graph-Bert}{\checkmark}}\\\hline
SE(3)-Transformer \cite{fuchs2020se}& & &\checkmark&\textcolor{blue}{\href{https://github.com/FabianFuchsML/se3-transformer-public}{\checkmark}}\\\hline
Mesh Graphormer~\cite{lin2021mesh} & \checkmark & \checkmark&  &\textcolor{blue}{\href{https://github. com/microsoft/MeshGraphormer}{\checkmark}}\\\hline
Gophormer~\cite{zhao2021gophormer} &  & &\checkmark  &\\\hline
EGT~\cite{hussain2021edge} & &\checkmark  &\checkmark  &\textcolor{blue}{\href{https://github.com/shamim-hussain/egt}{\checkmark}}\\\hline
SAN~\cite{kreuzer2021rethinking} &  & \checkmark& \checkmark&\textcolor{blue}{\href{https://github.com/DevinKreuzer/SAN}{\checkmark}}\\\hline
GraphiT~\cite{mialon2021graphit} & \checkmark &\checkmark &\checkmark &\textcolor{blue}{\href{https://github.com/inria-thoth/GraphiT}{\checkmark}}\\\hline
Graphormer~\cite{ying2021transformers} &  \checkmark &\checkmark & \checkmark &\textcolor{blue}{\href{https://github.com/microsoft/Graphormer}{\checkmark}}\\\hline
Mask-transformer \cite{min2022masked} &  & &\checkmark  &\textcolor{blue}{\href{https://github.com/qwerfdsaplking/F2R-HMT}{\checkmark}}\\\hline
 TorchMD-NET \cite{tholkeequivariant} &&&\checkmark&\textcolor{blue}{\href{https://github.com/torchmd/torchmd-net}{\checkmark}}\\

\bottomrule
\end{tabular}
}
}
\end{table}
\section{Transformer Architecture for Graphs}\label{sec.transformer_on_graphs}
The self-attention mechanism in the standard Transformer actually considers the input tokens as a fully-connected graph, which is agnostic to the intrinsic graph structure among the data. Existing methods that enable Transformer to be aware of topological structures are generally categorized into three groups: 1) GNNs as auxiliary modules in Transformer~(\textbf{GA}), 2) Improved positional embedding from graphs~(\textbf{PE}), 3) Improved attention matrices from graphs~(\textbf{AT}).  We summarize relevant literature in terms of these three dimensions in \cref{tab_summary}.

\begin{figure}[htbp]
\centering
\subfigure[Before Trans]{
\includegraphics[width=0.26\columnwidth]{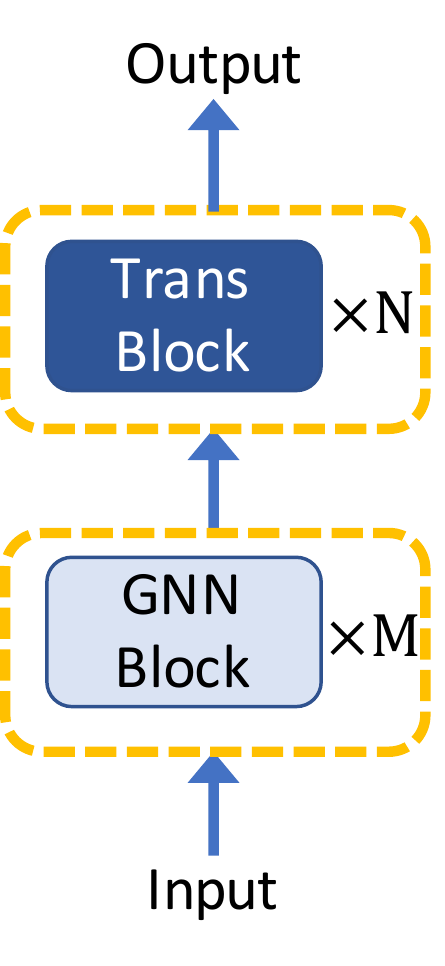}
}
\subfigure[Alternatively]{
\includegraphics[width=0.26\columnwidth]{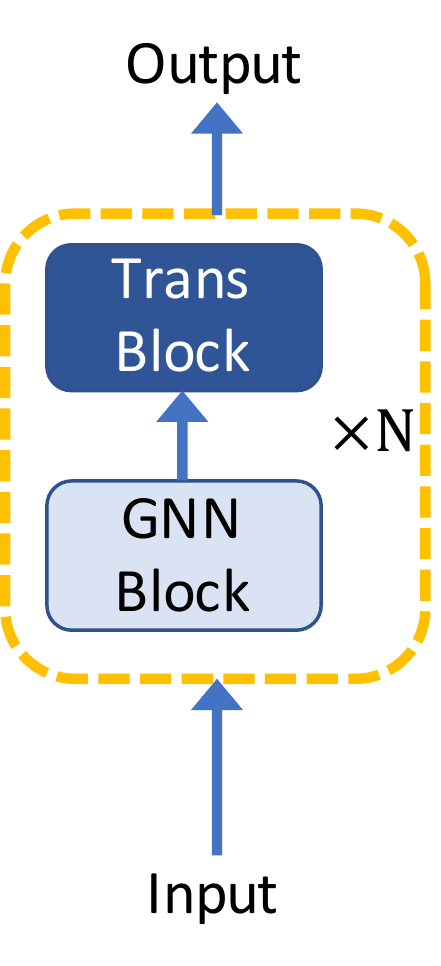}
}
\subfigure[Parallel]{
\includegraphics[width=0.39\columnwidth]{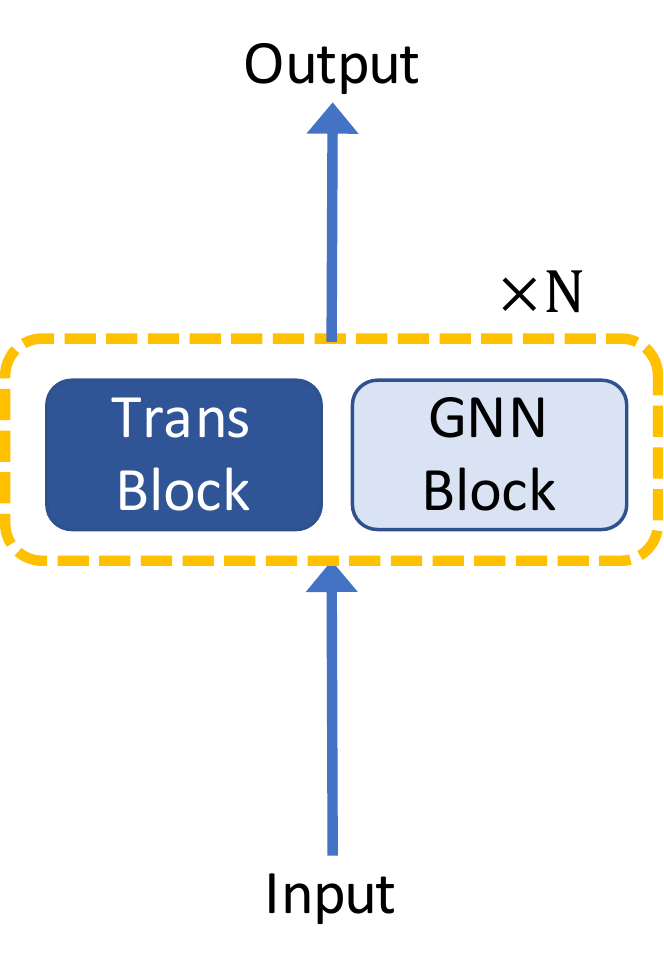}
}
\caption{Three types of GNN-as-Auxiliary-Modules architecture.}
\vspace{-2ex}
\label{gnn_emb}
\end{figure}

\subsection{GNNs as Auxiliary Modules in Transformer}
\label{sec.gnn_as_embedding}
The most direct solution of involving structural knowledge to benefit from global relation modeling of self-attention is to combine graph neural networks with Transformer architecture.
Generally, according to the relative postion between GNN layers and Transformer layers, existing Transformer architectures with GNNs are categorized into three types as illustrated in Figure \ref{gnn_emb}: (1) building Transformer blocks on top of GNN blocks, (2) alternately stacking GNN blocks and Transformer blocks, (3) parallelizing GNN blocks and Transformer blocks.

The first architecture is most-frequently adopted among the three options. For example,
GraphTrans \cite{jain2021representing} adds a Transformer subnetwork on top of a standard GNN layer. The GNN layer performs as a specialized architecture to learn local representations of the structure of a node's immediate neighbourhood, while the Transformer subnetwork computes all pairwise node interactions in a position-agnostic fashion, empowering the model global reasoning capability. GraphTrans is evaluated on graph classification task from biology, computer programming and chemistry, and achieves consistent improvement over benchmarks.

Grover \cite{rong2020self} consists of two GTransformer modules to represent node-level features and edge-level features respectively. In each GTransformer, the inputs are first fed into a tailored GNNs named dyMPN to extract vectors as queries, keys and values from nodes of the graph, followed by standard multi-head attention blocks. This bi-level information extraction framework enables the model to capture the structural information in molecular data and make it possible to extract global relations between nodes, enhancing the representational power of Grover.

GraphiT \cite{mialon2021graphit} also falls in the first architecture, which
adopts one Graph Convolutional Kernel Network (GCKN) \cite{chen2020convolutional} layer to produce a structure-aware representation from original features, and concatenate them as the input of Transformer architecture.
Here, GCKNs is a multi-layer model that produces a sequence of graph feature maps similar to a GNN. Different from GNNs, each layer of GCKNs enumerates local sub-structures at each node, encodes them using a kernel embedding, and aggregates the sub-structure representations as outputs. These representations in a feature map carry more structural information than traditional GNNs based on neighborhood aggregation.

Mesh Graphormer \cite{lin2021mesh} follows the second architecture by stacking a Graph Residual Block (GRB) on a multi-head self-attention layer as a Transformer block to model both local and global interactions among 3D mesh vertices and body joints. Specifically, given the contextualized features $\BM$ generated by multi-head self-attentions (MHSA) in Equation \ref{eq:attention3}, Mesh Graphormer improves the local interactions using a graph convolution in each Transformer block as:
\begin{equation}
\BM^{\prime}=\text{GraphConv}(\BA^G, \BM ; \BW^{G})=\sigma(\BA^G \BX \BW^{G}).
\end{equation}
where $\BA^G \in \mathbb{R}^{n \times n}$ denotes the adjacency matrix of a graph and $\mathbf{W}_{G}$ denotes the trainable parameters. $\sigma(\cdot)$ implies the non-linear activation function.
Mesh Graphormer also implements another two variants: building GRB before MHSA and building those two blocks in parallel, but they perform worse than the proposed one.

Graph-BERT \cite{zhang2020graph} adopts the third architecture by utilizing a graph residual term in each attention layer as follows:
\begin{equation}
    \BM'=\BM+\text{G-Res}(\BX,\BX_r,\BA^G),
\end{equation}
where the notation $\text { G-Res }(\BX, \BX_r,\BA^G)$ represents the graph residual term introduced in \cite{zhang2019gresnet} and $\BX_r$ is the raw features of all nodes in the graph.

\subsection{Improved Positional Embeddings from Graphs}
\label{pe_tf}
Although combining graph neural networks and Transformer has shown effectiveness in modeling graph-structured data, the best architecture to incorporate them remains an issue and requires heavy hype-parameter searching.
Therefore, it is meaningful to explore a graph-encoding strategy without adjustment of the Transformer architecture.
Similar to the positional encoding in Transformer for sequential data such as sentences, it is also possible to compress the graph structure into positional embedding (PE) vectors and add them to the input before it is fed to the actual Transformer model:
\begin{equation}
\widetilde{\BX} = \BX + f_\text{map}(\BP),
\end{equation}
where $\BX \in \mathrm{R}^{n \times d}$ is the matrix of input embeddings, $\BP \in \mathrm{R}^{n \times d_p}$ represents the graph embedding vectors, and $f_\text{map}: \mathrm{R}^{d_p} \rightarrow \mathrm{R}^{d}$ is a transformation network to align the dimension of both vectors.
The graph positional embedding $\BP$ is normally generated from the adjacent matrix $\BA^G \in \mathrm{R}^{n\times n}$.

\cite{dwivedi2020generalization} adopt Laplacian eigenvectors as $\BP$ in Graph Transformer. For each graph in the dataset, they pre-compute the Laplacian eigenvectors , which are defined by the factorization of the graph Laplacian matrix:
\begin{equation}
    \BU^{T} \Lambda \BU=\BI-\BD^{-1 / 2} \BA^G \BD^{-1 / 2},
\end{equation}
where $\BD$ is the degree matrix, and $\Lambda, \BU$ correspond to the eigenvalues and eigenvectors respectively. They use eigenvectors of the $k$ smallest non-trivial eigenvalues as the positional embedding, with $\BP \in \mathbb{R}^{n \times k}$ in this case. Since these eigenvectors have multiplicity occurring due to the arbitrary sign of eigenvectors, they randomly flip the sign of the eigenvectors during training.

\cite{hussain2021edge} employs pre-computed SVD vectors of the adjacent matrix as the positional embeddings $\BP$. They use the largest $r$ singular values and corresponding left and right singular vectors to represent the positional encoding. 

\begin{align}
&\BA^G \stackrel{\mathrm{SVD}}{\approx} \mathbf{U} \boldsymbol{\Sigma} \mathbf{V}^{T}=(\mathbf{U} \sqrt{\boldsymbol{\Sigma}}) \cdot(\mathbf{V} \sqrt{\boldsymbol{\Sigma}})^{T}=\hat{\mathbf{U}} \hat{\mathbf{V}}^{T}, \\
&\BP=\hat{\mathbf{U}} \| \hat{\mathbf{V}},
\end{align}
where $\BU,\BV \in \mathrm{R}^{n\times r}$ contain the $r$ left and right singular vectors corresponding to the top $r$ singular values in the diagonal matrix $\Sigma \in \mathbb{R}^{r\times r}$, $\|$ denotes concatenation operator along columns. They also randomly flip the signs during training as a form of data augmentation to avoid over-fitting, since similar to Laplacian eigenvectors, the signs of corresponding pairs of left and right singular vectors can be arbitrarily flipped. 

Different from Eigen PE and SVD PE, which attempt to compress the adjacent matrix into dense positional embeddings, there exist some heuristic methods that encode specific structural information from the extraction of the adjacent matrix.
For example, 
Graphormer \cite{ying2021transformers} uses the degree centrality as an additional signal to the neural network. To be specific, Graphformer assigns each node two real-valued embedding vectors according to its indegree and outdegree. For the $i$-th node, the degree-aware representation is denoted as:
\begin{equation}
    \tilde{\x} = \x + \z_{\mathrm{deg}^{-}(v_{i})}^{-}+\z_{\mathrm{deg}^{+}(v_{i})}^{+},
\end{equation}
where $\z^{-}, \z^{+} \in \mathbb{R}^{d}$ are learnable embedding vectors specified by the indegree $\operatorname{deg}^{-}(v_{i})$ and outdegree $\operatorname{deg}^{+}(v_{i})$ respectively. For undirected graphs, $\operatorname{deg}^{-}(v_{i})$ and $\operatorname{deg}^{+}(v_{i})$ could be unified to be $\operatorname{deg}(v_{i})$. By using the centrality encoding in the input, the softmax attention will catch the node importance signal in queries and keys of the Transformer.

Graph-BERT \cite{zhang2020graph} introduces three types of PE to embed the node position information, \emph{i.e.}, an absolute WL-PE which represents different codes labeled by the Weisfeiler-Lehman algorithm, an intimacy based PE and a hop based PE which are both variant to the sampled subgraphs.  
\cite{cai2020graph} applies graph transformer to tree-structured abstract meaning representation (AMR) graph. It adopts a distance embedding for each node by encoding the minimum distance from the root node as a flag of the importance of the corresponding concept in the whole-sentence semantics. \cite{kreuzer2021rethinking} proposes a learned positional encoding that can utilize the full Laplacian spectrum to learn the position of each node in a given graph.

\subsection{Improved Attention Matrices from Graphs}
\label{at_tf}
Although node positional embedding is a convenient practice to inject graph priors into Transformer architectures, the progress of compressing graph structure into fixed-sized vectors suffers from information loss, which might limit their effectiveness. For this sake, another group of works attempts to improve the attention matrix computation based on graph information:
\vspace{-1ex}
\begin{equation}
\vspace{-1ex}
    \begin{split}
        \BA =& f_\text{G\_att}(\BX,\BA^G,\BE;\BQ,\BK,\BW_1),\\
        \BM =& f_\text{M}(\BX,\BA,\BA^G,\BE;\BV,\BW_2),
    \end{split}
\end{equation}
where $\BX$ is the input features, $\BA^G$ is the adjacent matrix of the graph, $\BE \in \mathbb{R}^{n\times n\times d_e}$ is the edge features if available, $\BQ,\BK,\BV$ are the attention parameters, $\BW_1,\BW_2$ are extra graph encoding parameters. 

One line of models adapts self-attention mechanism to GNN-like architectures by restricting a node only attending to local node neighbours in the graph, which can be computationally formulated as an attention masking mechanism:
\vspace{-1ex}
\begin{equation}
\vspace{-1ex}
\label{maskattention}
    \BA = (\frac{1}{\sqrt{d}} \mathbf{X Q}(\mathbf{X}\mathbf{K})^{\top}) \odot \BA^{G},
\end{equation}
where $\BA^G \in \mathbb{R}^{n \times n}$ is the adjacent matrix of the graph. In \cite{dwivedi2020generalization}, $\BA^G_{ij}=1$ if there is an edge between $i$-th node and $j$-th node. Given the simple and generic nature of this architecture, they obtain competitive performance against standard GNNs on graph datasets. In order to encode edge features, they also extend Equation \ref{maskattention} as:
\vspace{-1ex}
\begin{equation}
      \BA = (\frac{1}{\sqrt{d}} \mathbf{X Q}(\mathbf{X}\mathbf{K})^{\top}) \odot \BA^{G} \odot \BE'\BW_E,  
\end{equation}
where $\BW_E \in \mathbb{R}^{d_e\times d}$ is the parameter matrix, $\BE'$ is the edge embedding matrix from the previous layer which is updated based on $\BA$, we do not elaborate these details for simplicity.

One possible extension of this practice is masking the attention matrices of different heads with different graph priors. In the original multi-head self-attention blocks, different attention heads implicitly attend to information from different representation subspaces of different nodes. While in this case, using the graph-masking mechanism to enforce the heads explicitly attend to different subspaces with graph priors further improves the model representative capability for graph data.
For example,
\cite{yao2020heterogeneous} computes the attention based on the extended Levi graph. Since Levi graph is a heterogeneous graph that contains different types of edges. They ﬁrst group all edge types into a single one to get a homogeneous subgraph referred to as connected subgraph. The connected subgraph is actually an undirected graph that contains the complete connected information in the original graph. Then they split the input graph into multiple subgraphs according to the edge types. Besides learning the directly connected relations, they introduce a fully-connected subgraph to learn the implicit relationships between indirectly connected nodes. Multiple adjacent matrices are assigned to different attention heads to learn a better representation for AMR task.
\cite{min2022masked} adopts a similar practice, which carefully designs four types of interaction graphs for modeling neighbourhood relations in CTR prediction task: induced subgraph, similarity graph, cross-neighbourhood graph, and complete graph. And they use the masking mechanism to encode these graph priors to improve neighbourhood representation.

GraphiT \cite{mialon2021graphit} extends the adjacent matrix to a kernel matrix, which is more flexible to encode various graph kernels. Besides, they use the same matrix for keys and queries following the recommendation of \cite{tsai2019transformer} to reduce parameters without hurting the performance in practice, and adopt a degree matrix to reduce the overwhelming influence of highly connected graph components.
The update equation can be formulated as:
\begin{equation}
\aligned
        \BA &= (\frac{1}{\sqrt{d}} \mathbf{X Q}(\mathbf{X}\mathbf{Q})^{\top}) \odot \BK_r,\\
        \widetilde{\BX} &= \text{SoftMax}(\BA)(\BX\BV),\\
        \BM &= \text{LayerNorm}(\BD^{-\frac{1}{2}}\widetilde{\BX}+\BX),
\endaligned
\end{equation}
where $\BD \in \mathbb{R}^{n\times n}$ is the diagonal matrix of node degrees, $\BK_r \in \mathbb{R}^{n\times n}$ is the kernel matrix on the graph, which is used as diffusion kernel and p-step random walk kernel.

Another line of models attempts to add soft graph bias to attention scores.
Graphormer \cite{ying2021transformers} proposes a novel Spatial Encoding mechanism. Concretely, they consider a distance function $\phi(v_i,v_j)$, which measures the spatial relation between nodes $v_i$ and $v_j$ in the graph. They select $\phi(v_i,v_j)$ as the shortest path distance (SPD) between $v_i$ and $v_j$. If they are not connected, the output of $\phi$ is set as a special value, \emph{i.e.}, $-1$. They assign each feasible value of $\phi$ a learnable scale parameter as a graph bias term. So the update rule is:
\begin{equation}
    \BA = (\frac{1}{\sqrt{d}} \mathbf{X Q}(\mathbf{X}\mathbf{K})^{\top}) + \BB^{s}.
\end{equation}
$\BB^{s}$ is the bias matrix, where $\BB^{s}_{ij}=b_{\phi(v_i,v_k)}$ is the learnable scalar indexed by $\phi(v_i,v_k)$, and shared across all layers. 
In order to handle graph structure with edge features, they also design an edge feature bias term. Specifically, for each ordered node pair $(v_i,v_j)$, they search (one of) the shortest path $\text{SP}_{ij}=(e_1,e_2,\ldots,e_N)$ from $v_i$ to $v_j$, and then compute an average of dot-products of the edge features and a learnable embedding along the path. Combined with the above spatial bias, the unnormalized attention score can be modified as:
\begin{equation}
    \BA = (\frac{1}{\sqrt{d}} \mathbf{X Q}(\mathbf{X}\mathbf{K})^{\top}) + \BB^{s}+\BB^{c}
\end{equation}
where $\BB^{c}$ is the edge feature bias matrix. $\BB^{c}_{ij}=\frac{1}{N}\sum^N_{n=1}x_{e_n}(w_n^E)^\top$, where $x_{e_{n}}$ is the feature of the $n$-th edge $e_{n}$ in $\mathrm{SP}_{i j}, w_{n}^{E} \in \mathbb{R}^{d_{e}}$ is the $n$-th weight embedding, and $d_{e}$ is the dimensionality of edge feature.

Gophormer \cite{zhao2021gophormer} proposes proximity-enhanced multi-head attention (PE-MHA) to encode multi-hop graph information. Specifically, for a node pair $\langle v_{i}, v_{j}\rangle$, $M$ views of structural information is encoded as a proximity encoding vector, denoted as  $\phi_{i j} \in \mathbb{R}^{M}$, to enhance the attention mechanism. The proximity-enhanced attention score $\BA_{i j}$ is defined as:
\vspace{-1.5ex}
\begin{equation}
\vspace{-1.5ex}
  \BA_{ij} =    (\frac{1}{\sqrt{d}} \x_i\BQ(\x_j\BK)^\top)+\boldsymbol{\phi}^{ij} \b^\top,
\end{equation}
where $\b \in \mathbb{R}^{M}$ is the learnable parameters that compute the bias of structural information.  The proximity encoding is calculated by $M$ structural encoding functions defined as:
\begin{equation}
\boldsymbol{\phi}_{i j}=\operatorname{Concat}(\Phi_{m}(v_{i}, v_{j}) \mid m \in 0,1, \ldots, M-1),
\end{equation}
where each structural encoding function $\Phi_{m}(\cdot)$
encodes a view of structural information. 

PLAN \cite{khoo2020interpretable} also proposes a structure aware self-attention to model the tree structure of rumour propagation in social media. The modified attention calculation can be defined as:
\vspace{-1.5ex}
\begin{align}
\vspace{-2.5ex}
  \BA_{ij} &=    \frac{1}{\sqrt{d}}( \x_i \BQ(\x_j\BK)^\top)+a_{ij}^K\\
  \BM_i &= \sum^n_{j=1}\text{SoftMax}(\BA_{ij})(\x_j\BV+a_{ij}^V)
\end{align}
Both $a_{ij}^{V}$ and $a_{ij}^{K}$ are learnable parameter vectors that represent one of the ﬁve possible structural relationships between the pair of the tweets (i.e. parent, child, before, after and self).


 \begin{table}[h]
\centering
\caption{Details of the datasets.}
\vspace{-2ex}
\resizebox{0.45\textwidth}{!}{
\scriptsize
\begin{tabular}{@{\hskip 0.03in}c@{\hskip 0.03in}|@{\hskip 0.03in}c@{\hskip 0.05in}c@{\hskip 0.05in}c@{\hskip 0.05in}c@{\hskip 0.05in}c@{\hskip 0.05in}c}
\toprule
Dataset Type             & Name         & \#Graph & \#Nodes(Avg.) & \#Edges(Avg.)  & Task type                  & Metric   \\ \midrule
\multirow{3}{*}{Graph-level} & ZINC         & 12,000  & 23.2    & 49.8     & Regression                 & MAE      \\ 
                      & ogbg-molhiv  & 41,127  & 25.5    & 27.5     & Binary cla.      & ROC-AUC  \\ 
                       & ogbg-molpcba & 437,929 & 26.0    & 28.1     & Binary cla.      & AP       \\ \midrule
\multirow{3}{*}{Node-level}  & Flickr       & 1       & 89,250   & 899,756   & Multi-class cla. & Accuracy \\ 
                       & ogbg-arxiv   & 1       & 169,343  & 1,166,243  & Multi-class cla. & Accuracy \\ 
                       & ogbg-product & 1       & 2,449,029 & 61,859,140 & Multi-class cla. & Accuracy \\ \bottomrule
\end{tabular}
}
\vspace{-2ex}
 \label{datasets}
\end{table}

\begin{table}[h]
\caption{Hyper-parameters of various Transformer sizes.}
\vspace{-2ex}
\centering
\setlength\tabcolsep{4pt}{
\resizebox{0.8\linewidth}{!}{
\begin{tabular}{llll}
\toprule
\multicolumn{1}{l|}{}                              & \multicolumn{3}{|c}{Transformer Size} \\
\multicolumn{1}{l|}{}                              & Small & Middle & Large \\ 
 \midrule
\multicolumn{1}{l|}{\#Layers}                      & 6     & 12     & 12    \\ 
\multicolumn{1}{l|}{Hidden Dimension}              & 80    & 80     & 512   \\ 
\multicolumn{1}{l|}{FFN Inner Hidden Dimension}    & 80    & 80     & 512   \\ 
\multicolumn{1}{l|}{\#Attention Heads}             & 8     & 8      & 32    \\ 
\multicolumn{1}{l|}{Hidden Dimension of Each Head} & 10    & 10     & 16    \\ 
\bottomrule  
\end{tabular}
}
}
\vspace{-2ex}
\label{hype-parameters}
\end{table}

\/*
\begin{table*}[h]
\footnotesize
\centering
\caption{Performance comparison on graph-level tasks. For each column, the numbers with background color are the best results in each group. The numbers in bold  are the best results across the groups.}
\begin{tabular}{cc|ccc|ccc|ccc}
\toprule
   &        & \multicolumn{3}{c|}{ZINC(MAE$\downarrow$)} & \multicolumn{3}{c|}{ogbg-molhiv(ROC-AUC$\uparrow$)} & \multicolumn{3}{c}{ogbg-molpcba(AP$\uparrow$)} \\
   &        &     Small & Middle &  Large &                Small & Middle &  Large &            Small & Middle &  Large \\
\midrule
TF & vanilla &    0.6689 & 0.6700 & 0.6699 &               0.7466 & 0.7230 & 0.7269 &           0.1624 & 0.1673 & 0.1676 \\\midrule
\multirow{3}{*}{GA} & \cellcolor{LightCyan} before &    0.4700 & 0.4809 & 0.5169 &              0.6758 & 0.7339 & 0.7182 &           0.2105 & 0.1989 &\cellcolor{LightCyan} 0.2269 \\
   &\cellcolor{LightCyan}  alter &    \cellcolor{LightCyan} 0.3771 & 0.3412 & \cellcolor{LightCyan} 0.2956 &              \cellcolor{LightCyan}  0.7200 &  0.7086 & \cellcolor{LightCyan} 0.7433 &           \cellcolor{LightCyan} 0.2474 & 0.2417 & 0.2244 \\
   &  \cellcolor{LightCyan}  parallel &    0.3803 & \cellcolor{LightCyan} \textbf{0.2749} & 0.2984 &               0.7138 &\cellcolor{LightCyan} 0.7750 & 0.7603 &           0.2345 & \cellcolor{LightCyan} 0.2444 & 0.2205 \\\midrule
\multirow{3}{*}{PE} & \cellcolor{LightYellow}  degree &    \cellcolor{LightYellow} 0.5364 & \cellcolor{LightYellow} 0.5364 & 0.5435 &               \cellcolor{LightYellow} 0.7506 & 0.6818 & \cellcolor{LightYellow} 0.7357 &           0.1672 & 0.1646 & 0.1650 \\
   &\cellcolor{LightYellow} eig &    0.6031 & 0.6074 & 0.6166 &               0.7407 & \cellcolor{LightYellow} 0.7279 & 0.7351 &           \cellcolor{LightYellow} 0.2194 &\cellcolor{LightYellow}  0.2087 & \cellcolor{LightYellow} 0.2131 \\
   & \cellcolor{LightYellow} svd &    0.5811 & 0.5462 & \cellcolor{LightYellow} 0.5400 &                0.7350 & 0.7155 & 0.7275 &           0.1648 & 0.1663 & 0.1767 \\\midrule
\multirow{4}{*}{AT} & \cellcolor{LightRed} SPB &    0.5122 & 0.4878 & 0.6100 &               0.7065 & 0.7589 & 0.7255 &           0.2409 & 0.2524 &  \cellcolor{LightRed} \textbf{0.2621} \\
   &\cellcolor{LightRed} PMA &    0.6194 & 0.6057 & 0.5963 &               0.6902 & 0.7054 & 0.7314 &           0.2115 & 0.1956 & 0.2518 \\
   &\cellcolor{LightRed} Mask-1 &   \cellcolor{LightRed} \textbf{0.2861} &  \cellcolor{LightRed} 0.2772 & \cellcolor{LightRed} \textbf{0.2894} &               \cellcolor{LightRed} \textbf{0.7610} &  \cellcolor{LightRed} \textbf{0.7753} &  \cellcolor{LightRed} \textbf{0.7960} &           0.2573 &  \cellcolor{LightRed} \textbf{0.2662} & 0.1594 \\
   & \cellcolor{LightRed} Mask-n &    0.3906 & 0.3596 & 0.4765 &               0.7286 & 0.7423 & 0.7128 &            \cellcolor{LightRed} \textbf{0.2619} & 0.2577 & 0.2380 \\
\bottomrule
\end{tabular}
\label{graph-level}
\end{table*}

\begin{table*}[h]
\footnotesize
\centering
\caption{Performance comparison on node-level tasks. For each column, the numbers with background color are the best results in each group. The numbers in bold  are the best results across the groups.}
\begin{tabular}{cc|ccc|ccc|ccc}
\toprule
   &        & \multicolumn{3}{c|}{Flickr(Acc$\uparrow$)} & \multicolumn{3}{c|}{ogbn-arxiv(Acc$\uparrow$)} & \multicolumn{3}{c}{ogbn-products(Acc$\uparrow$)} \\
   &        &       Small & Middle &  Large &           Small & Middle &  Large &              Small & Middle &  Large \\
\midrule
TF & vanilla &      0.5270 & 0.5279 & 0.5214 &          0.5539 & 0.5571 & 0.5598 &             0.7887 & 0.7887 & 0.7956 \\
\midrule
\multirow{3}{*}{GA} & \cellcolor{LightCyan} before &      0.5352 & 0.5369 &\cellcolor{LightCyan}0.5272 &          0.5608 & 0.5590 & \cellcolor{LightCyan} 0.5614 &              \cellcolor{LightCyan} \textbf{0.7953} & 0.7888 & 0.8012 \\
   &  \cellcolor{LightCyan} alter &       \cellcolor{LightCyan} \textbf{0.5374} & 0.5357 & 0.5162 &          0.5599 & 0.5555 & 0.5592 &             0.7905 & \cellcolor{LightCyan} 0.7915 &  \cellcolor{LightCyan} \textbf{0.8057} \\
   & \cellcolor{LightCyan} parallel &      0.5370 &  \cellcolor{LightCyan} \textbf{0.5379} & 0.5209 &           \cellcolor{LightCyan} \textbf{0.5647} & \cellcolor{LightCyan} 0.5600 & 0.5529 &             0.7878 & 0.7896 & 0.7999 \\
\midrule
\multirow{3}{*}{PE} & \cellcolor{LightYellow} degree &     \cellcolor{LightYellow} 0.5291 & 0.5250 & 0.5133 &          0.5551 & 0.5618 & 0.5502 &           \cellcolor{LightYellow}  0.7920 & \cellcolor{LightYellow} 0.7913 & 0.7947 \\
   & \cellcolor{LightYellow} eig &      0.5253 & 0.5278 & \cellcolor{LightYellow} 0.5257 &         \cellcolor{LightYellow} 0.5575 & 0.5637 & 0.5658 &             0.7893 & 0.7887 & \cellcolor{LightYellow} 0.8017 \\
   & \cellcolor{LightYellow} svd &      0.5281 & \cellcolor{LightYellow} 0.5317 & 0.5203 &          0.5614 &  \cellcolor{LightYellow}\textbf{0.5663} &  \cellcolor{LightYellow}\textbf{0.5706} &             0.7856 & 0.7893 & 0.8007 \\
\midrule
\multirow{4}{*}{AT} &\cellcolor{LightRed}  SPB &      \underline{0.5368} & \underline{0.5364} & 0.5234 &          0.5503 & \underline{0.5605} & 0.5576 &             0.7921 &  \underline{\textbf{0.7918}} & 0.7984 \\
   & \cellcolor{LightRed} PMA &      0.5240 & 0.5288 & 0.5204 &          0.5567 & 0.5571 & 0.5492 &             0.7877 & 0.7853 & 0.7945 \\
   &\cellcolor{LightRed}  Mask-1 &      0.5295 & 0.5300 & 0.5236 &          0.5598 & 0.5559 & 0.5583 &             0.7923 & 0.7917 & 0.7963 \\
   &\cellcolor{LightRed}  Mask-n &      0.5359 & 0.5349 &  \underline{\textbf{0.5348}} &          0.5593 & 0.5603 & 0.5576 &             \underline{0.7935} & 0.7917 & \underline{0.8016} \\
\bottomrule
\end{tabular}
\label{node-level}
\end{table*}
*/

\begin{table*}[h]
\scriptsize
\centering
\caption{Performance comparison on graph-level tasks. For each column, the values with underline are the best results in each group. The values in bold  are the best results across the groups.}
\vspace{-2ex}
\resizebox{1\linewidth}{!}{
\begin{tabular}{@{\hskip 0.03in}c@{\hskip 0.04in}c@{\hskip 0.03in}|@{\hskip 0.03in}c@{\hskip 0.05in}c@{\hskip 0.05in}c@{\hskip 0.03in}|@{\hskip 0.03in}c@{\hskip 0.05in}c@{\hskip 0.05in}c@{\hskip 0.03in}|@{\hskip 0.03in}c@{\hskip 0.05in}c@{\hskip 0.05in}c@{\hskip 0.03in}|@{\hskip 0.03in}c@{\hskip 0.05in}c@{\hskip 0.05in}c@{\hskip 0.03in}|@{\hskip 0.03in}c@{\hskip 0.05in}c@{\hskip 0.05in}c@{\hskip 0.03in}|@{\hskip 0.03in}c@{\hskip 0.05in}c@{\hskip 0.05in}c@{\hskip 0.03in}}
\toprule
   &        & \multicolumn{9}{c|@{\hskip 0.03in}}{ graph-level tasks } & \multicolumn{9}{c}{node-level tasks} \\
   &        & \multicolumn{3}{c|@{\hskip 0.03in}}{ZINC(MAE$\downarrow$)} & \multicolumn{3}{c|@{\hskip 0.03in}}{molhiv(ROC-AUC$\uparrow$)} & \multicolumn{3}{c|@{\hskip 0.03in}}{molpcba(AP$\uparrow$)}  & \multicolumn{3}{c|@{\hskip 0.03in}}{Flickr(Acc$\uparrow$)} & \multicolumn{3}{c|@{\hskip 0.03in}}{arixv(Acc$\uparrow$)} & \multicolumn{3}{c}{product(Acc$\uparrow$)} \\
   &        &     Small & Middle &  Large &                Small & Middle &  Large &            Small & Middle &  Large  &       Small & Middle &  Large &           Small & Middle &  Large &              Small & Middle &  Large \\
\midrule
TF & vanilla &    0.6689 & 0.6700 & 0.6699 &               0.7466 & 0.7230 & 0.7269 &           0.1624 & 0.1673 & 0.1676 &      0.5270 & 0.5279 & 0.5214 &          0.5539 & 0.5571 & 0.5598 &             0.7887 & 0.7887 & 0.7956 \\\midrule
\multirow{3}{*}{GA} &  before &    0.4700 & 0.4809 & 0.5169 &              0.6758 & 0.7339 & 0.7182 &           0.2105 & 0.1989 & \underline{0.2269} &      0.5352 & 0.5369 & \underline{0.5272} &          0.5608 & 0.5590 & \underline{0.5614} &             \underline{\textbf{0.7953}} & 0.7888 & 0.8012\\
   & alter &    \underline{0.3771} & 0.3412 & \underline{0.2956} &            \underline{0.7200} &  0.7086 & 0.7433 &            \underline{0.2474} & 0.2417 & 0.2244 &      \underline{\textbf{0.5374}} & 0.5357 & 0.5162 &          0.5599 & 0.5555 & 0.5592 &             0.7905 &  \underline{0.7915} &   \underline{\textbf{0.8057}} \\
   &  parallel &    0.3803 & \underline{\textbf{0.2749}} & 0.2984 &               0.7138 & \underline{0.7750} & \underline{0.7603} &           0.2345 & \underline{0.2444} & 0.2205  &      0.5370 &   \underline{\textbf{0.5379}} & 0.5209 &            \underline{\textbf{0.5647}} & \underline{0.5600} & 0.5529 &             0.7878 & 0.7896 & 0.7999 \\\midrule
\multirow{3}{*}{PE} &   degree &    \underline{0.5364} & \underline{0.5364} & 0.5435 &   \underline{0.7506} & 0.6818 & \underline{0.7357} &           0.1672 & 0.1646 & 0.1650 &     \underline{0.5291} & 0.5250 & 0.5133 &          0.5551 & 0.5618 & 0.5502 &  \underline{0.7920} & \underline{0.7913} & 0.7947 \\
   & eig &    0.6031 & 0.6074 & 0.6166 &               0.7407 & \underline{0.7279} & 0.7351 & \underline{0.2194} & \underline{0.2087} & \underline{0.2131} &      0.5253 & 0.5278 & \underline{0.5257} & 0.5575 & 0.5637 & 0.5658 &             0.7893 & 0.7887 &  \underline{0.8017} \\
   &  svd &    0.5811 & 0.5462 & \underline{0.5400} &                0.7350 & 0.7155 & 0.7275 &           0.1648 & 0.1663 & 0.1767 &      0.5281 & \underline{0.5317} & 0.5203 &          \underline{0.5614} &  \underline{\textbf{0.5663}} &  \underline{\textbf{0.5706}} &             0.7856 & 0.7893 & 0.8007 \\\midrule
\multirow{4}{*}{AT} & SPB &    0.5122 & 0.4878 & 0.6100 &               0.7065 & 0.7589 & 0.7255 &           0.2409 & 0.2524 &  \underline{\textbf{0.2621}} &     \underline{0.5368} & \underline{0.5364} & 0.5234 &          0.5503 & \underline{0.5605} & 0.5576 &             0.7921 &  \underline{\textbf{0.7918}} & 0.7984 \\
   & PMA &    0.6194 & 0.6057 & 0.5963 &               0.6902 & 0.7054 & 0.7314 &           0.2115 & 0.1956 & 0.2518 &      0.5240 & 0.5288 & 0.5204 &          0.5567 & 0.5571 & 0.5492 &             0.7877 & 0.7853 & 0.7945 \\
   & Mask-1 &   \underline{\textbf{0.2861}} &  \underline{0.2772} &  \underline{\textbf{0.2894}} &                \underline{\textbf{0.7610}} &  \underline{\textbf{0.7753}} &  \underline{\textbf{0.7960}} &           0.2573 & \underline{\textbf{0.2662}} & 0.1594  &      0.5295 & 0.5300 & 0.5236 &          \underline{0.5598} & 0.5559 & \underline{0.5583} &             0.7923 & 0.7917 & 0.7963 \\
   &  Mask-n &    0.3906 & 0.3596 & 0.4765 &               0.7286 & 0.7423 & 0.7128 &            \underline{\textbf{0.2619}} & 0.2577 & 0.2380 &      0.5359 & 0.5349 &  \underline{\textbf{0.5348}} &          0.5593 & 0.5603 & 0.5576 &             \underline{0.7935} & 0.7917 & \underline{0.8016} \\
\bottomrule
\end{tabular}
}
\vspace{-3ex}
\label{tab:all_res}
\end{table*}

\begin{table}[]
\footnotesize
    \centering
    \caption{The training hyper-parameters.}
    \vspace{-2ex}
    \resizebox{0.9\linewidth}{!}{
    \begin{tabular}{l r|l r}
    \hline
        attention dropout & 0.1 & FFN dropout & 0.1 \\
        maximum steps & 1e+6 & warm-up steps & 4e+4 \\
        peak learning rate & 2e-4 & batch size & 256 \\
        weight decay value & 1e-3 & lr decay & Linear \\
        gradient clip norm & 5.0 & Adam $\epsilon, \beta_1, \beta_2$& 1e-8, 0.9, 0.99 \\
    \hline
    \end{tabular}
    }
    \vspace{-2ex}
    \label{tab:exp:train:hyperpara}
\end{table}

\section{Experimental Evaluations}\label{sec.exp}
We conduct an extensive evaluation to study the effectiveness of different methods. On the basis of standard Transformer, methods in the three groups, auxiliary GNN modules (GA), positional embeddings (PE), and improved attention (AT) are compared.
Since most methods are composed of more than one graph-specific module and are trained with various tricks, it is difficult to evaluate the effectiveness of each module in a fair and independent way. 
In our experiments, we extract the representative modules of existing models and evaluate their performance individually. 
For GA methods, we compare the three architectures described in Section \ref{sec.gnn_as_embedding}. In both \textit{alternately} and \textit{parallel} settings, the GNNs and Transformer layers are combined before the FFN layers.
PE methods include degree embedding \cite{ying2021transformers}, Laplacian Eigenvectors \cite{dwivedi2020generalization} and SVD vectors \cite{hussain2021edge}. 
AT methods contain spatial bias (SPB) \cite{ying2021transformers}, proximity-enhanced multi-head attention (PMA) \cite{zhao2021gophormer}, attention masked with 1-hop adjacent matrix (Mask-1) \cite{dwivedi2020generalization}.
We also mask attention with different hops of the adjacent matrix, denoted as (Mask-n), inspired by the multi-head masking mechanisms in \cite{yao2020heterogeneous,min2022masked}.
 We evaluate the methods on both graph-level tasks on small graphs and node-level tasks on a single large graph. 
 The details of the tasks and datasets are listed in Table \ref{datasets}.

 \subsection{Settings and Implementation Details}
To ensure a consistent and fair evaluation, we fix the scale of the Transformer architecture in three levels: small, middle, and large, 
whose configurations are listed in  Table \ref{hype-parameters}.
The remaining hyper-parameters are fixed to their empirical value as shown in Table~\ref{tab:exp:train:hyperpara}.
For node-level tasks in large-scale graphs, we adopt the shadow $k$-hop sampling \cite{zeng2020deep} to generate a subgraph for a target node, to which the graph-aware modules are applied.  We calculate the required inputs of all modules based on the subgraph instead of the original graph because most of them are not computationally scalable to large graphs.
We set the maximum sampling hop to be $2$ and the maximum neighbourhood per node to be $10$. 
For PE methods, we select the embedding size from $\{3,4,5\}$. For GA methods, we select the GNN type from GCN \cite{kipf2016semi}, GAT \cite{velivckovic2017graph} and GIN \cite{xu2018powerful}.
Our learning framework is built on PyTorch 1.8, and PyTorch Geometric 2.0, and the code is public at \textcolor{blue}{\href{https://github.com/qwerfdsaplking/Graph-Trans}{https://github.com/qwerfdsaplking/Graph-Trans}}.

\subsection{Experimental Results}
Table \ref{tab:all_res} reports the performance of ten graph-specific modules from three categories on six datasets of graph-level and node-level tasks, respectively. 
%
We summarize the main observations as follows:
\begin{itemize}[leftmargin=*]
\item Not surprisingly, in most cases, the evaluated graph-specific modules of Transformer lead to to better performance. For example, on molpcba, we observe at most a 56\% performance improvement compared with the vanilla Transformer.  This observation confirms the effectiveness of graph-specific modules on the various kinds of graph tasks. 
\item The improvement on graph-level tasks is more significant than that on node-level tasks. This may be due to the graph sampling process when dealing with a single large graph. The sampled induced subgraphs fail to keep the original graph intact, incurring variance and information loss in the learning process.
\item GA and AT methods bring more benefits than PE methods. 
In conclusion, PE's weaknesses are twofold.
First, as introduced in Section \ref{pe_tf}, PE does not contain intact graph information. Second, since PE is only fed into the input layer of the network, the graph-structural information would decay layer by layer across the model, leading to a degeneration of the performance.
\item It seems that different kinds of graph tasks enjoy different group of models. GA methods achieve the best performance in more than half of the cases (5 out of 9 cases) of node-level tasks. More significantly,  AT methods achieve the best performance in almost all the cases (8 out of 9 cases) of graph-level tasks. We conjecture that GA methods are able to better encode the local information of the sampled induced subgraphs in node-level tasks, while AT methods are suitable for modeling the global information of the single graphs in graph-level tasks. 
\end{itemize}
In summary, our experiments confirm the value of the existing graph-specific module on Transformer and reveal their advantages for a variety of graph tasks.

\section{Conclusion and Future Directions}\label{sec.conclusion}
In this survey, we present a comprehensive review of the current progress of applying Transformer model on graphs. 
Based on the injection part of graph data and graph model in Transformer, we classify the existing graph-specific modules into three typical groups: GNNs as auxiliary modules, improved positional embeddings from graphs, and improved attention matrices from graphs. 
To investigate the real performance gains under a fair 
setup, we implement the representative modules from three groups and compare them on six benchmarks with different tasks. 
We hope our systematic review and comparison will foster understanding and stimulate new ideas in this area.

In spite of the current successes, we believe several directions would be promising to investigate further and to start from in the future, including:
\begin{itemize}[leftmargin=*]
    \item \textbf{New paradigm of incorporating the graph and the Transformer.} 
    Most studies treat the graphs as strong prior to modifying the Transformer model. There is a great interest to develop the new paradigm that not just takes graphs as a prior, but also better reflects the properties of graphs. 
    \item \textbf{Extending to other kinds of graphs.} 
    Existing Graph Transformer models mostly focus on homogeneous graphs, which consider the node type and edge type as the same. It is also important to explore their potential on other forms of graphs, such as heterogeneous graphs and hypergraphs.
    \item \textbf{Extending to large-scale graphs.} Most existing methods are designed for small graphs, which might be computationally infeasible for large graphs. 
    As illustrated in our experiments, directly applying them to the sampled subgraphs would impair performance. Therefore, designing salable Graph-Transformer architecture is essential.
\end{itemize}

\clearpage
\bibliographystyle{named}
\bibliography{ijcai22}

\end{document}